\documentclass[3p,sort&compress,final]{elsarticle}
\usepackage{amsmath,amssymb,amsfonts,amsthm}
\usepackage[linesnumbered, ruled,vlined]{algorithm2e}
\SetAlFnt{\small}
\SetAlCapFnt{\small}
\SetAlCapNameFnt{\small}
\usepackage{graphicx,booktabs}
\usepackage{setspace}
\usepackage{color}
\usepackage{subcaption}
\usepackage{graphicx} 
\usepackage{float}  
\usepackage{chngcntr}

\usepackage[colorlinks=true]{hyperref}

\begin{document}
	
\begin{frontmatter}
	\title{GARCH-FIS: A Hybrid Forecasting Model with Dynamic Volatility-Driven  Parameter Adaptation}
	\author{Wen-Jing Li}
		\author{Da-Qing Zhang\corref{cor1}}
	\ead{d.q.zhang@ustl.edu.cn}
	\cortext[cor1]{Corresponding author}	

\address{School of Science, University of Science and Technology Liaoning,
	Anshan, Liaoning Province, 114051, PR China}

\begin{abstract}
This paper proposes a novel hybrid model, termed GARCH-FIS, for recursive  rolling multi-step forecasting of financial time series. 
It integrates a Fuzzy Inference System (FIS) with a Generalized Autoregressive Conditional Heteroskedasticity (GARCH) model to jointly address nonlinear dynamics and time-varying volatility. The core innovation is a dynamic parameter adaptation mechanism for the FIS, specifically activated within the multi-step forecasting cycle. In this process, the conditional volatility estimated by a rolling window GARCH model is continuously translated into a price volatility measure. At each forecasting step, this measure, alongside the updated mean of the sliding window data---which now incorporates the most recent predicted price---jointly determines the parameters of the FIS membership functions for the next prediction. Consequently, the granularity of the fuzzy inference adapts as the forecast horizon extends: membership functions are automatically widened during high-volatility market regimes to bolster robustness and narrowed during stable periods to enhance precision. This constitutes a fundamental advancement over a static one-step-ahead prediction setup. Furthermore, the model’s fuzzy rule base is automatically constructed from data using the Wang-Mendel method, promoting interpretability and adaptability. Empirical evaluation, focused exclusively on multi-step forecasting performance across ten diverse financial assets, demonstrates that the proposed GARCH-FIS model significantly outperforms benchmark models --- including Support Vector Regression(SVR), Long Short-Term Memory networks(LSTM), and an ARIMA-GARCH econometric model --- in terms of predictive accuracy and stability, while effectively mitigating error accumulation in extended recursive forecasts.
\end{abstract}
\end{frontmatter}

	\section{Introduction}

	In financial forecasting, traditional time series models have been widely employed. Among them, the Generalized Autoregressive Conditional Heteroskedasticity (GARCH) model and its variants have become mainstream tools for volatility modeling, primarily because they effectively capture volatility clustering \cite{engle1982, bollerslev1986}. On the other hand, the Fuzzy Inference System (FIS) offers unique advantages in handling uncertainty and imprecise information. By using linguistic rules to describe nonlinear relationships, the FIS can integrate expert knowledge with historical data, providing a viable approach for modeling complex financial systems \cite{fahmy2015}.
	
	Nevertheless, each approach has inherent limitations. GARCH-family models, while effective for volatility tracking, are grounded in a linear framework and offer limited capacity to model nonlinear structures and uncertainty directly. Their predictive performance often deteriorates during trend reversals and extreme market conditions. Conversely, conventional fuzzy systems typically rely on fixed membership function parameters, lacking a dynamic link to key risk factors such as market volatility. This makes them prone to overfitting during high volatility regimes and insufficiently responsive during calm periods \cite{varshney2023}, thereby constraining their wider adoption in financial forecasting.
	
To overcome these complementary shortcomings, this paper proposes a  hybrid forecasting model that integrates a FIS with a GARCH model. 
The integration is designed to leverage the strengths of both components: the GARCH(1,1) model captures the volatility clustering inherent in return series, while the FIS handles nonlinear mappings. The core innovation is most prominent in the recursive rolling  multi-step forecasting process. Here, the GARCH-estimated conditional volatility is transformed into a price volatility measure. Crucially, at each step of the multi-step prediction cycle, this measure, along with the updated mean of the sliding window data (which incorporates the latest forecast), dynamically determines the parameters of the FIS membership functions. This enables the model to adapt its inference granularity as the forecast horizon extends: widening the membership functions in high-volatility periods to enhance robustness and narrowing them in low-volatility phases to sharpen predictive resolution. This continuous adaptation during multi-step forecasting stands in contrast to a one-step-ahead prediction setup, where the FIS parameters, once determined from a historical window, remain fixed. Furthermore, the model automates the
 construction of its fuzzy rule base using the Wang-Mendel  (WM) method 
 \cite{wang1992,wang2003}, improving interpretability and self-adaptation.

The main contributions of this study are twofold:

\begin{enumerate}
	\item We establish a theoretical framework for coupling a FIS with a GARCH model within a recursive  rolling  multi-step forecasting setup. This clarifies the synergistic mechanism that combines the nonlinear mapping capability of the FIS with the volatility modeling strength of GARCH, and specifically details how the FIS parameters become adaptive during the multi-step prediction cycle.

	\item We propose an automatic rule base generation mechanism based on the WM method and a volatility-driven dynamic parameter adjustment strategy that operates throughout the multi-step forecasting process. This enables the fuzzy system to anticipate market regime changes and self-optimize its inference mechanism at each forecast step.
\end{enumerate}
	
	The remainder of the paper is structured as follows. Section 2 reviews related work on GARCH models, fuzzy inference systems, and their hybrid applications in finance. Section 3 details the proposed GARCH-FIS methodology, including the integrated framework, parameter identification, and the recursive rolling prediction procedure. Section 4 presents experimental results and a comparative analysis with benchmark models on multiple financial datasets. Finally, Section 5 concludes the paper and suggests directions for future research.

\section{Related Work}

Financial time series analysis has long centered on two core challenges: modeling time-varying volatility and capturing nonlinear dynamics. This section reviews relevant literature through this lens, examining first the GARCH family of models for volatility, then fuzzy inference systems for nonlinear forecasting, and finally hybrid approaches that attempt to integrate both.

\subsection{GARCH Models and Volatility Forecasting}

Volatility modeling remains a central theme in financial econometrics. The GARCH model, introduced by Bollerslev  as an extension of Engle's  ARCH framework, rapidly became the benchmark for capturing volatility clustering---the phenomenon where periods of high volatility tend to cluster together \cite{poon2003,engle2001}. The standard GARCH(1,1) formulation is:
\begin{equation}\label{eq:1}
	\begin{aligned}
	r_t &= \mu + \varepsilon_t, \quad \varepsilon_t \mid \mathcal{F}_{t-1} \sim \mathcal{N}(0, \sigma_t^2), \\
	\sigma_t^2 &= \omega + \alpha \varepsilon_{t-1}^2 + \beta \sigma_{t-1}^2,
\end{aligned}
\end{equation}
where \( r_t \) is the return, \( \varepsilon_t \) is the shock, and \( \sigma_t^2 \) is the conditional variance. The ARCH term (\( \alpha \)) captures the short-term impact of news, while the GARCH term (\( \beta \)) reflects volatility persistence. The model's success in forecasting short-term volatility for equity indices, exchange rates, and commodities is well-documented \cite{awartani2005, yang2006, hou2012}. Furthermore, it serves as a foundational tool for Value-at-Risk measurement, option pricing, and hedging strategy construction \cite{angelidis2004, lehar2002, chang2013}.

To address specific market features, numerous GARCH extensions have been developed. The EGARCH model uses a logarithmic variance equation to capture the asymmetric response of volatility to positive versus negative shocks \cite{nelson1991}. The TGARCH model introduces a dummy variable to differentiate the impact of good and bad news \cite{glosten1993}. The GARCH-M model incorporates volatility into the mean equation to account for risk premiums \cite{bollerslev1994}.

Despite their flexibility, GARCH-family models possess inherent limitations. They are fundamentally linear in variance and do not directly model the price level. More critically, their capacity to handle complex nonlinear structures and inherent uncertainty in financial series is limited. Consequently, their forecasting performance often deteriorates significantly during market regime shifts, trend reversals, or extreme events \cite{silvennoinen2009}.

\subsection{Fuzzy Inference Systems in Financial Forecasting}

Fuzzy Inference Systems (FIS), grounded in Zadeh's fuzzy set theory \cite{zadeh1965}, offer a distinct approach by modeling nonlinear relationships through linguistic rules, making them particularly suitable for addressing uncertainty and imprecise data. The Takagi-Sugeno-Kang (TSK) type FIS has become predominant in financial forecasting due to its output being a precise value, thus avoiding complex defuzzification.

A zero-order TSK FIS operates as follows: In puts are fuzzified via membership functions (e.g., triangular functions). A rule base of "IF-THEN" statements (e.g., IF \(x_1\) is \(A_1\) AND \(x_2\) is \(A_2\) THEN \(y = d_k\)) encodes knowledge. The final output is a weighted average of rule consequents (\(d_k\)), with weights determined by the firing strength of each rule.

Early research sought to enhance FIS learning. Jang  integrated the TSK system into an Adaptive Neuro-Fuzzy Inference System (ANFIS), using a hybrid learning algorithm to optimize parameters \cite{jang1993}. While powerful, ANFIS often requires manual setup of the initial rule structure and can converge to local optima. Subsequent studies applied FIS directly to forecasting: Hiemstra  built an interactive fuzzy logic system for the S\&P 500 \cite{hiemstra1994}; Kodogiannis \& Lolis  showed adaptive fuzzy systems outperformed pure neural networks in exchange rate prediction \cite{kodogiannis2002}; and Yang  used subtractive clustering for automatic rule generation from OHLC data \cite{yang2007}.

Recent advancements continue to refine FIS applications. Chang \& Liu  used TSK rules to capture stock price trends, outperforming linear models \cite{chang2008}. Esfahanipour \& Aghamiri  combined stepwise regression, FCM clustering, and ANFIS to achieve low MAPE for index prediction \cite{esfahanipour2010}. For exchange rates, Korol  incorporated fundamental variables into fuzzy rules \cite{korol2014}, while Maciel \& Ballini  proposed an interval-valued FIS for high-low price range forecasting \cite{maciel2021}. Other innovations include hybrid models combining fuzzy time series with ARFIMA and Particle Swarm Optimization \cite{sadaei2016}, big data frameworks for fuzzy forecasting \cite{wang2018}, and multivariate fuzzy models optimized with PSO \cite{bilal2024}.

However, a key limitation persists in traditional FIS applications: the parameters of membership functions are typically static. This lack of a dynamic link to market states, such as volatility, makes standard fuzzy systems prone to overfitting during turbulent periods and insufficiently sensitive during calm markets\cite{varshney2023}.

\subsection{Hybrid Models: Integrating GARCH and Fuzzy Systems}

Given the complementary strengths and limitations of GARCH and FIS, researchers have increasingly explored hybrid models to simultaneously capture volatility clustering and nonlinear dynamics.

One strand of work focuses on using fuzzy logic to enhance GARCH-type models. Hung  proposed an adaptive fuzzy GARCH model that characterized leverage and clustering effects via IF-THEN rules, with parameters estimated by a genetic algorithm \cite{hung2011}. Others have integrated fuzzy systems with GARCH variants; for example, Mohammed et al.  combined FIS, neural networks, and EGARCH using differential evolution to capture clustering and leverage effects \cite{mohamad2020}. Alternatively, some studies use GARCH to preprocess data for a fuzzy forecaster. Ibrahim et al.  used FARIMA-GARCH residuals as inputs to a fuzzy system, reporting lower error than a standalone FARIMA model \cite{ibrahim2021}. A conceptually different approach is found in Set-GARCH models, which use random set theory to establish a transformation from fuzzy volatility to actual volatility \cite{dai2025}.

While these hybrid approaches demonstrate promise, they often exhibit one or more of the following limitations: (1) reliance on expert knowledge or metaheuristic optimization for rule generation, which can be subjective and computationally intensive; (2) a lack of a direct, mechanistic link where GARCH-derived volatility dynamically regulates the FIS structure in real-time; and (3) limited use of fully data-driven methods for creating a parsimonious and interpretable rule base.

\subsection{Positioning of This Study}

This study proposes a hybrid forecasting model based on the GARCH-FIS architecture, aiming to integrate the strengths of both GARCH and fuzzy inference systems. The GARCH(1,1) component captures volatility clustering in return series, while the FIS handles nonlinear mappings. The core innovation lies in its recursive rolling  multi-step forecasting process, where fuzzy membership function parameters of the FIS are dynamically adjusted based on anticipated market conditions to enhance the model's out-of-sample long-term forecast accuracy.

\section{GARCH-FIS Model}

\subsection{Overall Framework of the Proposed Method}

The GARCH-FIS hybrid forecasting model developed in this study integrates a FIS for nonlinear mapping with a GARCH(1,1) model for dynamic parameter adaptation. The primary objective is to forecast the price \(P_{t+1}\), \(P_{t+2}\), \(\cdots\), \(P_{t+n}\) of a financial asset at time \(t+1, t+2,\cdots,t+n\), respectively, utilizing all available information up to time \(t\).

Given a historical price series \(\{P_t\}_{t=1}^T\), the core mathematical formulation of the model is:
\begin{equation}\label{eq:2}
	\hat{P}_{t+1} = \text{FIS}(X_t; \Theta_t),
\end{equation}

where \(\hat{P}_{t+1}\) is the predicted price, \(\text{FIS}(\cdot)\) represents the complete fuzzy inference process, and \(X_t\) is the input feature vector constructed from a rolling window of length \(W\): \(X_t = (P_{t-W+1}, \ldots, P_{t-1}, P_t)^T\). \(\Theta_t\) is the the dynamic parameter set, which encapsulates the adaptive capability of the model:
\begin{equation}\label{eq:3}
	\Theta_t = \{ (p_c(s), \hat{\sigma}_t(s)) \mid s = t-W+1, \ldots, t \},
\end{equation}
where \(p_c(s)\) is the center parameter and \(\hat{\sigma}_t(s)\) is the width parameter of the triangular membership functions for the data point at time \(s\) within the window. Crucially, the width parameter \(\hat{\sigma}_t\) is derived from the conditional volatility estimated by the GARCH(1,1) model, establishing a direct link between the forecasted market volatility and the model's fuzzification mechanism.

The GARCH(1,1) process models the return series \(\{r_t\}\), where \(r_t = (P_t - P_{t-1}) / P_{t-1} \times 100\):
\begin{equation}\label{eq:4}
		\varepsilon_t = r_t - \mu,
\end{equation}
\begin{equation}\label{eq:5}
		\sigma_t^2 = \omega + \alpha \varepsilon_{t-1}^2 + \beta \sigma_{t-1}^2.
	\end{equation} 
where, \(\varepsilon_t\) is the residual, \(\mu\) is the conditional mean, and \(\sigma_t^2\) is the conditional variance capturing volatility clustering.

To integrate this volatility measure into the price-forecasting framework, the conditional volatility \(\sigma_t\) is converted into a price volatility \(\hat{\sigma}_t\) that is dimensionally consistent with the price level:
\begin{equation}\label{eq:6}
	\hat{\sigma}_t = \bar{X}_t \times \frac{\sigma_t}{100},
\end{equation}
where \(\bar{X}_t = \frac{1}{W} \sum_{s=t-W+1}^{t} P_s\) is the average price within the rolling window, serving as a scaling benchmark.

\begin{figure}
	\begin{center}
		\includegraphics[width=1\textwidth]{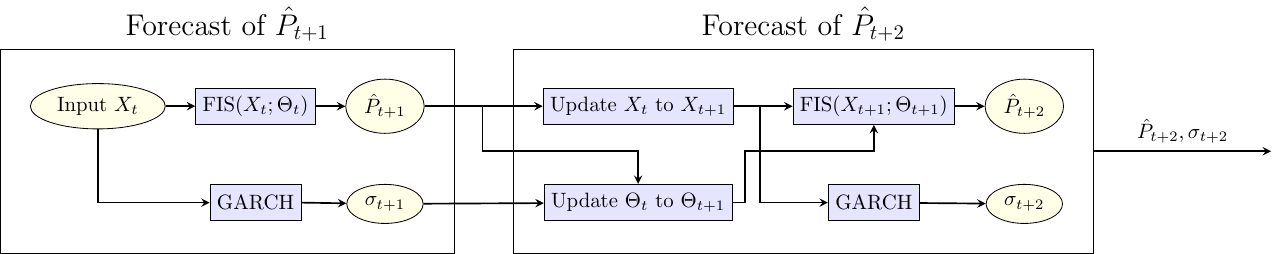}
		\caption{Recursive Rolling Forecasting Process of GARCH-FIS.
		}\label{fig:1}
	\end{center}
\end{figure}

Figure \ref{fig:1} illustrates the recursive rolling adaptive forecasting mechanism based on GARCH volatility estimation and a fixed-structure  FIS. Throughout the forecasting process, the rule set and universe partitioning of the FIS remain unchanged; only the parameters of the antecedent membership functions---including the universe size and function shape---are dynamically adjusted to adapt to market fluctuations.

The process begins at time \( t \) by feeding the current market feature vector \( X_t \) into the FIS parameterized with \( \Theta_t \). Through nonlinear fuzzy inference, the system generates the one-step-ahead price forecast \( \hat{P}_{t+1} \). Subsequently, using historical return data up to time \( t \), the GARCH(1,1) model estimates the conditional volatility for the next period, \( \sigma_{t+1} \), which is then transformed into price volatility \( \hat{\sigma}_{t+1} \).

Both the price forecast \( \hat{P}_{t+1} \) and the volatility estimate \( \hat{\sigma}_{t+1} \) are then embedded into the updated FIS parameter set \( \Theta_{t+1} \). This integration dynamically adjusts the width and universe coverage of the membership functions for the next forecasting cycle, enabling adaptive responsiveness.

The rolling window is then recursively updated by incorporating the newly predicted price \( \hat{P}_{t+1} \) into the data window. At the next time step \( t+1 \), the updated FIS with parameters \( \Theta_{t+1} \) processes the new feature vector \( X_{t+1} \) to produce the subsequent forecast \( \hat{P}_{t+2} \). This sequence repeats iteratively, resulting in a continuous recursive rolling adaptive prediction loop.

Overall, the mechanism embodies a closed-loop adaptive cycle of  forecast  $\to$ volatility estimation $\to$ parameter and data updating $\to$ re-forecast, maintaining a stable FIS structure while modulating its parameters via predicted prices and estimated volatility to dynamically align with evolving market conditions.

In summary, the model employs a recursive rolling forecasting mechanism that integrates volatility modeling and nonlinear mapping. The GARCH-derived volatility ${\hat{\sigma}}_{t+1}$ together with the price forecast ${\hat{P}}_{t+1}$ is embedded as an internal regulatory variable within the parameter set \(\Theta_{t+1} \) of the fuzzy system. Since the FIS structure---including its rule base and universe partition---remains fixed, adaptation is achieved solely through dynamic adjustment of membership function parameters. This design enables the framework to simultaneously capture nonlinear price trends and autonomously adapt to time-varying market volatility. As a result, the system continuously updates both its forecasting parameters and data window in a closed-loop manner, making it particularly suited for forecasting in complex and non-stationary financial environments.

\subsection{Parameter Identification}

This section details the identification methods for two parameter categories:  GARCH model parameters and FIS's  parameters and rule base.

 \subsubsection{GARCH Model Parameter Identification}
 
 \begin{algorithm}
 	\caption{GARCH Model Parameter Estimation}
 	\label{alg:garch_estimation}
 	\SetAlgoLined
 	\KwIn{Data of Window: win = $\{P_{t-W+1}, P_{t-W+2}, \ldots, P_t\}$}
 	\KwOut{$\mu, \sigma^2_{t+1}$}
 	
 	$\{r_t\} \leftarrow \text{Compute\_Returns}(\text{win})$ \tcp*{via Eq. \eqref{eq:4}}
 	$\theta = (\mu, \omega, \alpha, \beta) \leftarrow \text{FitGARCH}(\{r_t\})$ \tcp*{via Eq. \eqref{eq:8}}
 	$\sigma^2_{t+1} \leftarrow \text{Forecast\_Conditional\_Variance}(\theta, \{r_t\})$ \tcp*{via Eq. \eqref{eq:5}}
 	\Return $\mu, \sigma^2_{t+1}$
 \end{algorithm}

Within the hybrid framework, the GARCH(1,1) model serves to characterize the time-varying conditional heteroskedasticity of returns and to translate this into a price-volatility measure for adaptive control of the FIS. We estimate the GARCH parameters using a rolling window Maximum Likelihood Estimation (MLE) strategy. This approach dynamically updates parameter estimates to capture short-term volatility evolution, ensuring temporal consistency with the forecasting framework, in contrast to traditional static MLE which fails to adapt to changing market structures.

Let \( \theta = (\mu, \omega, \alpha, \beta) \) denote the parameter vector of the GARCH(1,1) model. Given a rolling window of length \( W \), the log-likelihood function for the return samples within the window \( \{r_i\}_{i=t-W+1}^{t} \) is:
\begin{equation}\label{eq:7}
	\mathcal{L}_t(\theta) = -\frac{1}{2} \sum_{i=t-W+1}^{t} \left( \ln(2\pi) + \ln \sigma_i^2 + \frac{\varepsilon_i^2}{\sigma_i^2} \right),
\end{equation}
where \( \varepsilon_i = r_i - \mu \) and \( \sigma_i^2 = \omega + \alpha \varepsilon_{i-1}^2 + \beta \sigma_{i-1}^2 \). Parameter estimation is achieved by maximizing this likelihood subject to standard constraints ensuring positivity and stationarity:
\begin{equation}\label{eq:8}
	\begin{aligned}
	& \underset{\theta}{\text{maximize}} && \mathcal{L}_t(\theta) \\
	& \text{subject to} && \omega > 0,\ \alpha \geq 0,\ \beta \geq 0,\ \alpha + \beta < 1.
\end{aligned}
\end{equation}

We solve this constrained optimization problem using the quasi-Newton BFGS algorithm \cite{byrd1995}, which efficiently handles the nonlinear likelihood maximization and allows dynamic tracking of parameter evolution within the rolling window.

\subsubsection{Fuzzy Inference System Parameter Identification}

Our identification framework is data-driven, employing price mean and price volatility to configure membership function parameters---specifically their centers and widths---and leveraging the WM method for automatic rule generation.

(1) {\it Offline Training: }Construction of Initial FIS Parameters

In the training phase, the complete FIS structure---including the antecedent fuzzy sets, the rule base, and the consequent parameters---is identified once using a dedicated historical dataset. Assume we have historical prices \(P_1, \ldots, P_T\) and a sliding window length \(W\). For each time \(t\) (\(W \leq t \leq T\)), we define the training window as \(\{P_{t-W+1}, \ldots, P_t\}\). The parameters for each window are calculated as follows.

The sample mean \(\bar{X}_t\) of the window is computed as:
\begin{equation}\label{eq:9}
	\bar{X}_t = \frac{1}{W} \sum_{k=t-W+1}^{t} P_k,
\end{equation}
which corresponds to the \(W\)-day moving average at time \(t\)---a common technical indicator for short-term trend analysis in financial price series. The price volatility \(\hat{\sigma}_t\) is taken as the standard deviation of prices within the same window:
\begin{equation}\label{eq:10}
	\hat{\sigma}_t = \sqrt{\frac{1}{W-1} \sum_{k=t-W+1}^{t} \left(P_k - \bar{X}_t\right)^2}.
\end{equation}

The centers of the membership functions are distributed symmetrically around \(\bar{X}_t\), using \(\hat{\sigma}_t\) as the spacing interval. For each input variable, the five center points are given by:
\begin{equation}\label{eq:11}
	p^{k}_{c,j}=\bar{X}_t+(j-3)\hat{\sigma}_t,\quad j=1,\ldots,5,
\end{equation}
yielding centers at \(\bar{X}_t-2\hat{\sigma}_t\), \(\bar{X}_t-\hat{\sigma}_t\), \(\bar{X}_t\), \(\bar{X}_t+\hat{\sigma}_t\), and \(\bar{X}_t+2\hat{\sigma}_t\). These centers correspond to price levels that are far below, below, near, above, and far above the moving average, respectively, providing a fuzzy quantification of the price position relative to the prevailing trend.

The half-width of all triangular membership functions is set equal to \(\hat{\sigma}_t\). Thus, for a center \(p_{c,j}\), the support interval is \([p_{c,j}-\hat{\sigma}_t, p_{c,j}+\hat{\sigma}_t]\), with the membership function defined as:
\begin{equation}\label{eq:12}
	\mu_{A_j}(y;p_{c,j},\hat{\sigma}_t)=\begin{cases}
	\dfrac{\hat{\sigma}_t-|y-p_{c,j}|}{\hat{\sigma}_t}, & y\in[p_{c,j}-\hat{\sigma}_t, p_{c,j}+\hat{\sigma}_t],\\
	0, & \text{otherwise}.
\end{cases}
\end{equation}

The five membership functions constructed by \eqref{eq:10}-\eqref{eq:12} thus represent, in fuzzy terms, the distance of the price from the moving average at time \(t\): “far below MA”, “below MA”, “near MA”, “above MA”, and “far above MA”. This establishes a direct, interpretable link between volatility-scaled fuzzy partitions and conventional technical analysis concepts.

\begin{algorithm}
	\caption{Offline Training of FIS using WM Method (Part I) }
	\label{alg:fis_training_fixed}
	\DontPrintSemicolon
	\KwIn{Historical price series $\{P_t\}_{t=1}^T$, Window size $W$, Forecast horizon $h$ (e.g., $h=1$)}
	\KwOut{FIS with fixed rule base and antecedent fuzzy sets, along with the parameter set $\Theta$ for the last window}
	
	Initialize global parameter dictionary $\Theta \gets \{\}$ \tcp*{Store membership function parameters (centers and widths) for each time step}
	
	Initialize candidate rules set $R_{\text{candidates}} \gets\{ \varnothing\}$\;
	
	\For{$t = W$ \textbf{to} $T-h$}{
		$\text{win} \gets \{P_{t-W+1}, \dots, P_t\}$ \tcp*{Current sliding window}
		
		$y_{\text{target}} \gets P_{t+h}$ \tcp*{Target value for prediction}
		
		$\mu_\text{w} \gets \text{mean}(\text{win})$\;
		$\sigma_\text{w} \gets \text{std}(\text{win})$\;
		\If{$\sigma_\text{w} = 0$}{
			$\sigma_\text{w} \gets 1.0$\;
		}
		
		$\text{centers} \gets [\mu_\text{w} - 2\sigma_\text{w}, \mu_\text{w} - \sigma_\text{w}, \mu_\text{w}, \mu_\text{w} + \sigma_\text{w}, \mu_\text{w} + 2\sigma_\text{w}]$ \tcp*{According to Eq. \eqref{eq:11}}
		
		\tcp{Initialize or update the membership function parameters for the current time step $t$}
		\If{$t$ not in $\Theta$}{
			$\Theta[t] \gets (\text{centers}, \sigma_\text{w})$ \tcp*{First time as the window end, use current window's $\sigma_\text{w}$ as the width}
		}
		
		\tcp{For historical time steps in the window ($t-W+1$ to $t-1$), if not initialized, set width to 1.0 (only occurs in the first window)}
		\For{$s = t-W+1$ \textbf{to} $t-1$}{
			\If{$s$ not in $\Theta$}{
				$\Theta[s] \gets (\text{centers}, 1.0)$ \tcp*{Use current window's centers, set width to initial value 1.0}
			}
		}
		
		\tcp{Construct rule antecedents using the stored parameters for each time step in the window}
		$\text{antecedent} \gets [\ ]$\;
		\For{$s = t-W+1$ \textbf{to} $t$}{
			$(\text{centers}_s, \text{width}_s) \gets \Theta[s]$\;
			$x_s \gets P_s$ \tcp*{Current price as the feature}
			$\text{memberships} \gets \text{compute\_membership}(x_s, \text{centers}_s, \text{width}_s)$\;
			$\text{best\_label} \gets \arg\max(\text{memberships})$\;
			$\text{antecedent.append}(\text{best\_label})$\;
		}
	
		$R_{\text{candidates}}\text{.append}((\text{antecedent},\; y_{\text{target}}))$\;
	}
\end{algorithm}

\setcounter{algocf}{1}

\begin{algorithm}
	\caption{Offline Training of FIS using WM Method (Part II) }

\setcounter{AlgoLine}{23}
	\tcp{Merge duplicate antecedents and compute weighted average consequents}
	$R_{\text{final}} \gets \{\varnothing\}$\;
	\For{each unique antecedent $A$ in $R_{\text{candidates}}$}{
		$\text{consequent\_samples} \gets [y \text{ for } (\text{ant}, y) \text{ in } R_{\text{candidates}} \text{ if } \text{ant} = A]$\;
		$d_k \gets \text{weighted\_average}(\text{consequent\_samples})$ \tcp*{According to Eq. \eqref{eq:13}}
		$R_{\text{final}}\text{.add}(\text{Rule: IF } A \text{ THEN } y = d_k)$\;
	}
	
	\tcp{Extract the parameter set for the last window (used for initialization in forecasting)}
	$\Theta_{\text{last}} \gets \{\varnothing\}$\;
	\For{$s = (T-h)-W+1$ \textbf{to} $T-h$}{
		$\Theta_{\text{last}}[s] \gets \Theta[s]$\;
	}
	
	$\text{FIS} \gets \text{initialize\_fis}(\text{rule\_set}=R_{\text{final}}, \text{param\_dict}=\Theta_{\text{last}})$\;
	\Return FIS with  $\Theta_{\text{last}}$\;
\end{algorithm}

(2) {\it  Offline Training: }Rule Consequent Parameter Identification

We employ the WM method and its weighted-average enhancement to
 automatically generate the fuzzy rule base from the training data. 

 For each unique antecedent, the consequent value \(d_k\) is a weighted average of all training outputs \(y_l\), weighted by the sample’s activation strength \(\lambda_l(A^{})\) for that rule:
\begin{equation}\label{eq:13}
	d_k=\frac{\sum_{l=1}^{L}\lambda_l(A^{})\cdot y_l}{\sum_{l=1}^{L}\lambda_l(A^{})}, \quad \text{with } \lambda_l(A^{})=\prod_{i=1}^{n}\mu_{A^{}_i}(x_{l,i}),
\end{equation}
where \(\mu_{A^{}_i}(x_{l,i})\) is the membership degree of the \(l\)-th training sample to the fuzzy set \(A^{}_i\) in the rule antecedent. Each unique antecedent is paired with its calculated consequent to form a complete fuzzy rule.  Algorithm \ref{alg:fis_training_fixed}  summarizes the modeling process of the FIS in detail.

\subsection{Recursive Rolling Forecasting with Adaptive Parameter Adjustment  }

During the multi‑step recursive rolling forecasting process, the rule base  of the FIS remains fixed. Adaptation to evolving market conditions is achieved solely through dynamic adjustment of the membership‑function parameters (centers and widths) in each rolling step. This creates a direct link between the forecasted market volatility and model granularity: wider functions during high‑volatility periods enhance robustness, while narrower functions during low‑volatility periods improve sensitivity to subtle price changes.

The system is designed to emulate a real‑time forecasting environment, where the prediction for time \(t+1\) uses only information available up to time \(t\), aligning with practical trading‑decision scenarios.

Single‑step forecast:  
At each time \(t\), the one-step-ahead forecast is generated by the FIS as shown in equation \eqref{eq:2}. This forecasting process utilizes the feature vector \(X_t\) derived from the current data window and a time-specific parameter set \(\Theta_t\) identified specifically from information available up to time \(t\), 
where \(\Theta_t\) encodes the membership‑function parameters that modulate the FIS’s granularity according to the most recent volatility estimate and price level.

Multi‑step recursive rolling forecast (horizon \(n>1\)):  
The system proceeds through a sequence of recursive rolling steps. Each step produces a forecast and updates the parameters and data window for the next iteration:

(1) Forecast: Compute \(\hat{P}_{t+1}\) using \(X_t\) and \(\Theta_t\).  

(2) Estimate volatility: Apply the GARCH(1,1) model to data window at  time \(t\) to obtain the conditional volatility estimate \(\sigma_{t+1}\) and its price‑transformed version \(\hat{\sigma}_{t+1}\).  

(3) Update parameters: Construct the next parameter set \(\Theta_{t+1}\) by embedding the forecast \(\hat{P}_{t+1}\) and the volatility estimate \(\hat{\sigma}_{t+1}\). The update formulas are functionally identical to those used in training: the centers are given by \(p_{c,j}^{k} = \bar{X}_{t+i} + (j-3)\hat{\sigma}_{t+i}\) and the half‑width is set to \(\hat{\sigma}_{t+i}\).  

(4) Roll the window: Form the new input vector \(X_{t+1}\) by incorporating \(\hat{P}_{t+1}\) and shifting the window forward, effectively treating the forecast as an observed datum for the next recursion.  

(5) Repeat: With the updated \(X_{t+1}\) and \(\Theta_{t+1}\), compute \(\hat{P}_{t+2}\). Repeat steps (1)-(4) until the desired horizon \(n\) is reached.

The general recursive rolling relation for step \(i\;(1\leq i\leq n)\) can be compactly expressed as:
\begin{equation}\label{eq:15}
\begin{split}
		\Theta_{t+i-1} &= g\big(\mathcal{I}_{t+i-2}\big), \\
	X_{t+i-1} &= \big[\hat{P}_{t+i-1},\hat{P}_{t+i-2},\ldots,P_{t-W+i}\big], \\
	\hat{P}_{t+i} &= \text{FIS}\big(X_{t+i-1};\Theta_{t+i-1}\big),
\end{split}
\end{equation} 
where \(\mathcal{I}_{t+i-2}\) denotes all information available at time \(t+i-2\), \(g(\cdot)\) represents the parameter‑identification procedure, and for \(i=1\) we set \(\hat{P}_{t}=P_{t}\).

\begin{figure}
	\begin{center}
		\includegraphics[width=1\textwidth]{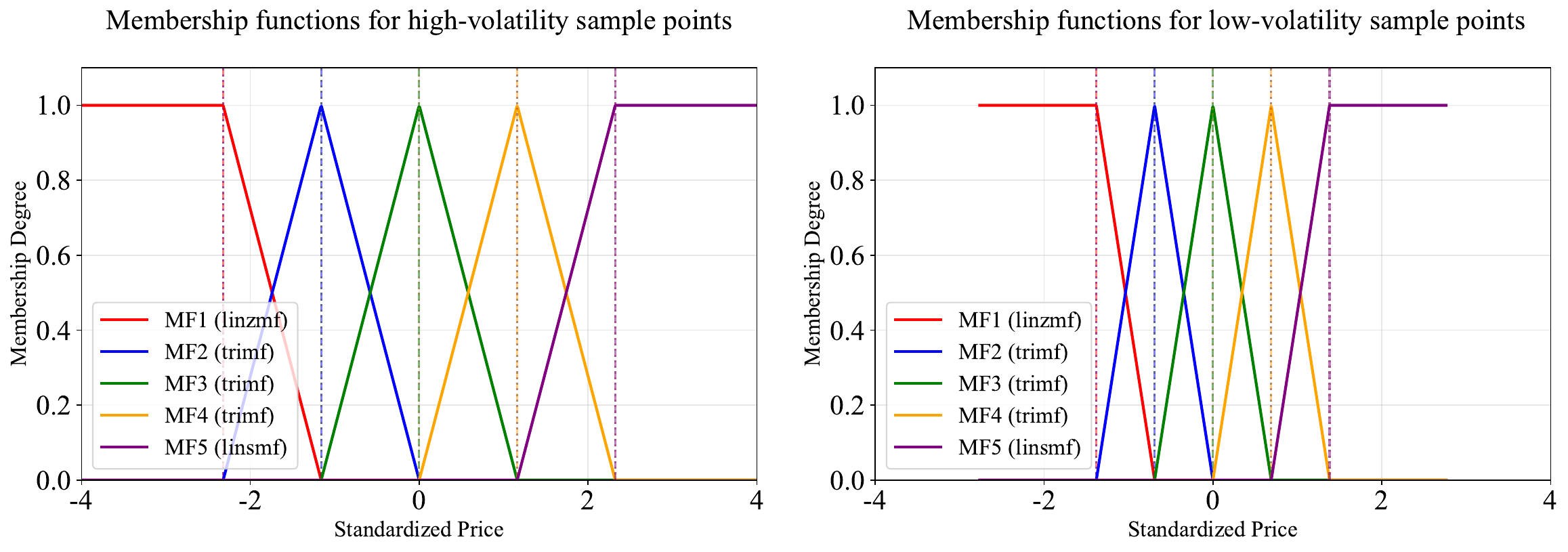}
		\caption{Distribution diagram of membership function for sample points during high/low fluctuation periods
		}\label{fig:2}
	\end{center}
\end{figure}

The rule consequents, identified during training, 
are not recalculated---preserving the system’s learned mapping 
while allowing its input interface to adapt dynamically. 
The membership-function parameters are updated using the latest price-volatility estimate \(\hat{\sigma}_{t+i}\) and the current window mean price \(\bar{X}_{t+i}\). 
This linear scaling of the membership-function support width with 
the volatility estimate is visually confirmed in Figure \ref{fig:2}, 
which compares membership‑function distributions for sample points from high- and low‑volatility periods.

	\begin{algorithm}[htbp]
	\caption{GARCH-FIS Multi-step Prediction}
	\label{alg:garch_fis_multi_step}
	\SetAlgoLined
	\KwIn{win = $\{P_{t-W+1}, P_{t-W+2}, \ldots, P_t\}$, FIS(win; $\Theta$)}
	\KwOut{Prediction sequence $P = [\hat{P}_{t+1}, \hat{P}_{t+2}, \hat{P}_{t+3}, \ldots, \hat{P}_{t+n}]$}
	
	$P \gets \{\varnothing\}$\;
	\For{$i = 1$ \KwTo $n$}{
		$\hat{P}_{t+i} \leftarrow \text{FIS}(\text{win}; \Theta)$ \tcp*{Predict next price}
		$(\sigma^2, \mu) \leftarrow \text{Algorithm \ref{alg:garch_estimation}}(\text{win})$ \tcp*{Update GARCH parameters}
		Update win: drop the first value and append $\hat{P}_{t+i}$ \tcp*{Shift window}
		$\bar{X}_{t+i} \leftarrow \text{mean}(\text{win})$\;
		$\hat{\sigma} \leftarrow \bar{X}_{t+i} \times \frac{\sigma}{\sqrt{100}}$ \tcp*{Price volatility via Eq.\eqref{eq:6}}
		Update $\Theta$ with $\hat{\sigma}_{t+i} \leftarrow \hat{\sigma}$ and $\hat{P}_{t+i}$ \tcp*{Update FIS parameters}
		Append $\hat{P}_{t+i}$ to $P$\;
	}
	\Return $P$
\end{algorithm}

The resulting closed‑loop system continuously regenerates adaptive FIS parameters and rolls the data window forward with each new forecast. By integrating its own forecasted path and the latest volatility information, the model achieves robust multi‑step predictions while strictly adhering to the non‑leakage principle (using only past and present data). The framework is thus both adaptive to time‑varying market conditions and structurally stable across successive rolling windows.

\section{Experiments and Results Analysis}

\subsection{Data Selection and Preprocessing}

To evaluate the generalization capability of the proposed GARCH-FIS model across diverse market environments and asset types, we conduct experiments on a comprehensive dataset comprising the daily closing prices of ten financial products from January 1, 2015, to December 31, 2024. The selected assets span major financial markets (China, US, Japan) and include key classes such as stock indices (e.g., CSI 300, S\&P 500), commodity futures (gold, copper), foreign exchange rates (USD\/JPY, EUR\/USD), fixed income, and individual equities. This variety ensures a robust test of the model's applicability.

A strict chronological split is applied: the first 80\% of the data is used for model training and the remaining 20\% constitutes the test set. This partitioning preserves the temporal order of financial time series, prevents look-ahead bias, and provides a sufficiently long test period to assess forecasting stability.

To ensure stable operation of the FIS and facilitate cross-asset comparison, input features are standardized within the training set using Z-score normalization:
\begin{equation}\label{eq:16}
	X = \frac{X_{\text{raw}} - \mu_{X_{\text{raw}}}}{\sigma_{X_{\text{raw}}}},
\end{equation}
where \(X_{\text{raw}}\) is the raw price, and \(\mu_{X_{\text{raw}}}\) and \(\sigma_{X_{\text{raw}}}\) are the  sample mean and standard deviation, respectively. This step removes scale differences and enhances the adaptability of the fuzzy rules.

The rolling window length \(W\) balances model responsiveness to market changes with estimation stability. An overly small \(W\) leads to noisy parameter estimates, while an overly large \(W\) weakens adaptability to structural breaks. We systematically evaluated a candidate set \(W \in \{3, 5, 10, 15\}\) trading days using grid search based on MAE, MAPE, and \(R^2\). Empirical results across all datasets indicated optimal performance over \(W = 10\) trading days (approximately two weeks). This length aligns with the concept of short-term trends in technical analysis, suggesting it effectively captures relevant market dynamics.

\subsection{Evaluation Metrics and Benchmark Models}

Three complementary evaluation metrics are employed to provide a comprehensive assessment of model performance: Mean Absolute Error (MAE), Mean Absolute Percentage Error (MAPE), and the Coefficient of Determination ($R^2$)\cite{srivastava2024association,makridou2013gold}. These metrics collectively measure the absolute deviation, relative error, and explanatory power of the forecasts, respectively.

Mean Absolute Error (MAE): Quantifies the average magnitude of absolute forecast errors:
\begin{equation}\label{eq:17}
	\text{MAE} = \frac{1}{N} \sum_{t=1}^{N} |\hat{y}_t - y_t|.
\end{equation}

Mean Absolute Percentage Error (MAPE): Expresses the forecast error as a percentage relative to the actual values, facilitating cross-scale comparison:
\begin{equation}\label{eq:18}
	\text{MAPE} = \frac{100\%}{N} \sum_{t=1}^{N} \left| \frac{\hat{y}_t - y_t}{y_t} \right|.
\end{equation}

Coefficient of Determination ($R^2$): Measures the proportion of variance in the actual data explained by the model:
\begin{equation}\label{eq:19}
	R^2 = 1 - \frac{\sum_{t=1}^{N} (\hat{y}_t - y_t)^2}{\sum_{t=1}^{N} (y_t - \bar{y})^2}.
\end{equation}
where, \(N\) is the number of forecast samples, \(\hat{y}_t\) is the predicted value, \(y_t\) is the actual observed value, and \(\bar{y}\) is the sample mean of the actual values.

To ensure a rigorous and fair comparison, we benchmark the proposed GARCH-FIS model against three mainstream forecasting approaches: Support Vector Regression (SVR), Long Short-Term Memory networks (LSTM), and an ARIMA-GARCH econometric model. The parameters for all benchmark models were systematically optimized to ensure they performed at their best capacity:

   Support Vector Regression (SVR): We employed a grid search to optimize the regularization parameter \(C\), the insensitivity loss parameter \(\varepsilon\), and the kernel coefficient \(\gamma\), selecting the combination that minimized the mean absolute error under cross-validation.
   
   Long Short-Term Memory (LSTM): Given its high training complexity, a rapid evaluation strategy was adopted. We tested predefined hyperparameter combinations covering the number of neurons, dropout rate, and training epochs, selecting the conuration that yielded the lowest MAE on a validation set.
   
   ARIMA-GARCH Econometric Model: The ARIMA orders \((p, d, q)\) were automatically selected from a predefined range based on the Akaike Information Criterion (AIC), coupled with a standard GARCH(1,1) specification to capture volatility clustering.

This systematic parameter tuning guarantees that the benchmark models contribute to the evaluation with their optimal performance, thereby validating the comparability of the results and the robustness of our conclusions.

\subsection{Experimental Results and Analysis}

\begin{figure}[H]
	\begin{center}
		\includegraphics[width=\textwidth]{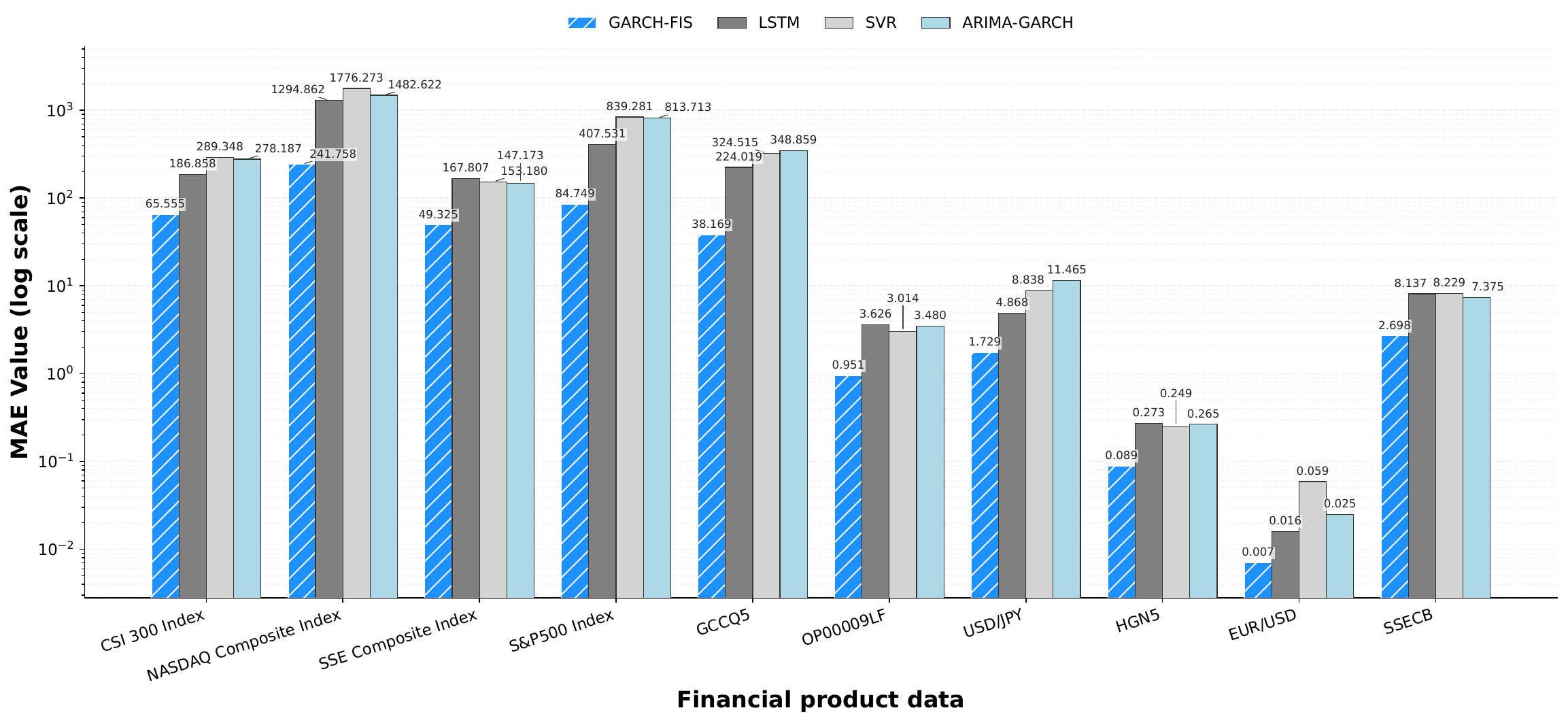}
		\caption{Comparison bar chart of MAE for different models across various financial products
		}\label{fig:3}
	\end{center}
\end{figure}

\begin{figure}
	\begin{center}
		\includegraphics[width=\textwidth]{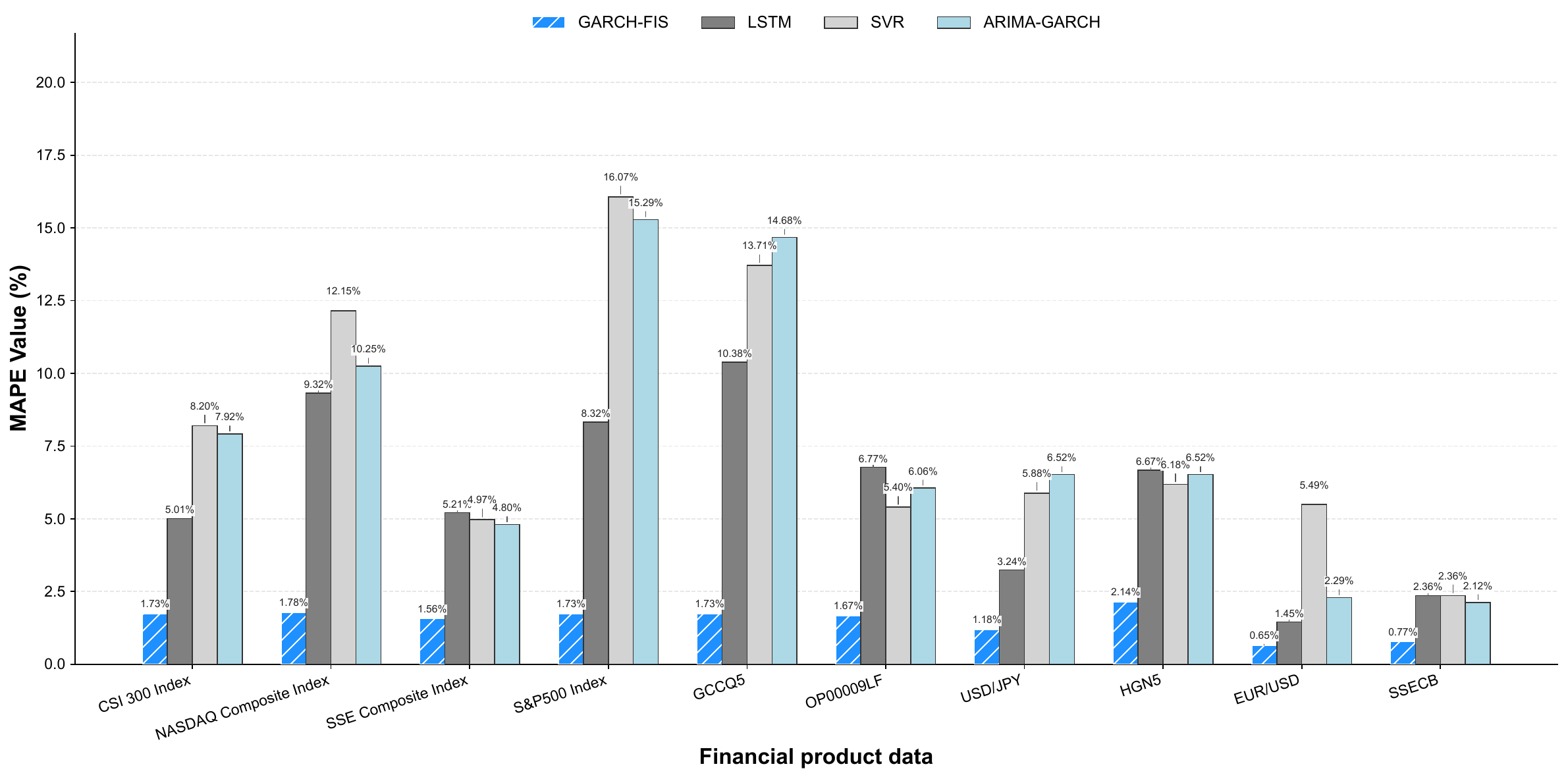}
		\caption{Comparison bar chart of MAPE for different models across various financial products
		}\label{fig:4}
	\end{center}
\end{figure}

The comparative forecasting performance of the proposed GARCH-FIS model against the three benchmark models across all ten financial datasets is systematically summarized in Figure \ref{fig:3} , Figure \ref{fig:4} and Table \ref{tab:r2_results}. The results consistently demonstrate the superior accuracy and robustness of the GARCH-FIS approach.

\begin{table}[htbp]
	\centering\small
	\caption{Comparison of Determination Coefficient (\(R^2\)) Results}
	\label{tab:r2_results}
	\begin{tabular}{lcccc}
		\toprule
		Financial product data & LSTM & SVR & ARIMA-GARCH & GARCH-FIS \\
		\midrule
		CSI 300 Index & 0.319 & -0.653 & -0.575 & \textbf{0.812} \\
		NASDAQ Composite Index & 0.064 & -1.105 & -0.402 & \textbf{0.965} \\
		SSE Composite Index & -0.765 & -0.108 & -0.025 & \textbf{0.743} \\
		S\&P500 Index & 0.454 & -1.400 & -1.576 & \textbf{0.973} \\
		GCQ5 & 0.159 & -1.304 & -1.592 & \textbf{0.965} \\
		0P000090LF & -0.151 & 0.261 & -0.111 & \textbf{0.893} \\
		USD/JPY & 0.443 & -0.653 & -2.326 & \textbf{0.930} \\
		HGN5 & -0.146 & 0.068 & -0.009 & \textbf{0.840} \\
		EUR/USD & -0.409 & -19.474 & -1.748 & \textbf{0.693} \\
		SSECB & -0.335 & -0.279 & -0.101 & \textbf{0.774} \\
		\bottomrule
	\end{tabular}
\end{table}

As shown in Figure \ref{fig:3}, the GARCH-FIS model achieves the lowest MAE for every asset, indicating its superior precision in minimizing absolute forecast deviations. The advantage is substantial. For instance, on the CSI 300 Index, the MAE of GARCH-FIS (65.555) is approximately 65\%, 77\%, and 76\% lower than those of LSTM, SVR, and ARIMA-GARCH, respectively. This pattern of significant error reduction holds across assets with diverse volatility profiles, such as Gold Futures (GCQ5), where GARCH-FIS attains an MAE of 38.169, markedly outperforming the benchmarks.

The relative error metric, MAPE, presented in Figure \ref{fig:4}, further confirms the accuracy of GARCH-FIS. Its MAPE values are consistently below 2\% for most assets, substantially lower than those of the benchmark models. For example, its MAPE for the CSI 300 Index is 1.73\%, compared to 5.01\%-8.20\% for the benchmarks. Even for less volatile assets like the Shanghai Convertible Bond Index (SSECB), GARCH-FIS maintains a low MAPE of 0.77\%, demonstrating effective error control in calm market regimes.

The $R^2$ results in Table \ref{tab:r2_results} provide the most compelling evidence of the model's explanatory power. The GARCH-FIS model yields high, positive  $R^2$ values across all datasets (e.g., 0.965 for NASDAQ, 0.973 for S\&P 500), signifying an excellent fit to the underlying data dynamics. In stark contrast, the benchmark models show unstable and often negative  $R^2$ values (e.g., -2.326 for ARIMA-GARCH on USD/JPY), indicating that their forecasts can be less accurate than a simple mean prediction. This stark difference underscores the limitation of traditional models in capturing the complex dynamics of financial series and highlights the effective integration of nonlinear mapping and volatility adaptation in our proposed framework.

\begin{figure}[htbp]
	\centering 
\begin{subfigure}[b]{0.48\textwidth}
	\centering
	\includegraphics[width=\textwidth]{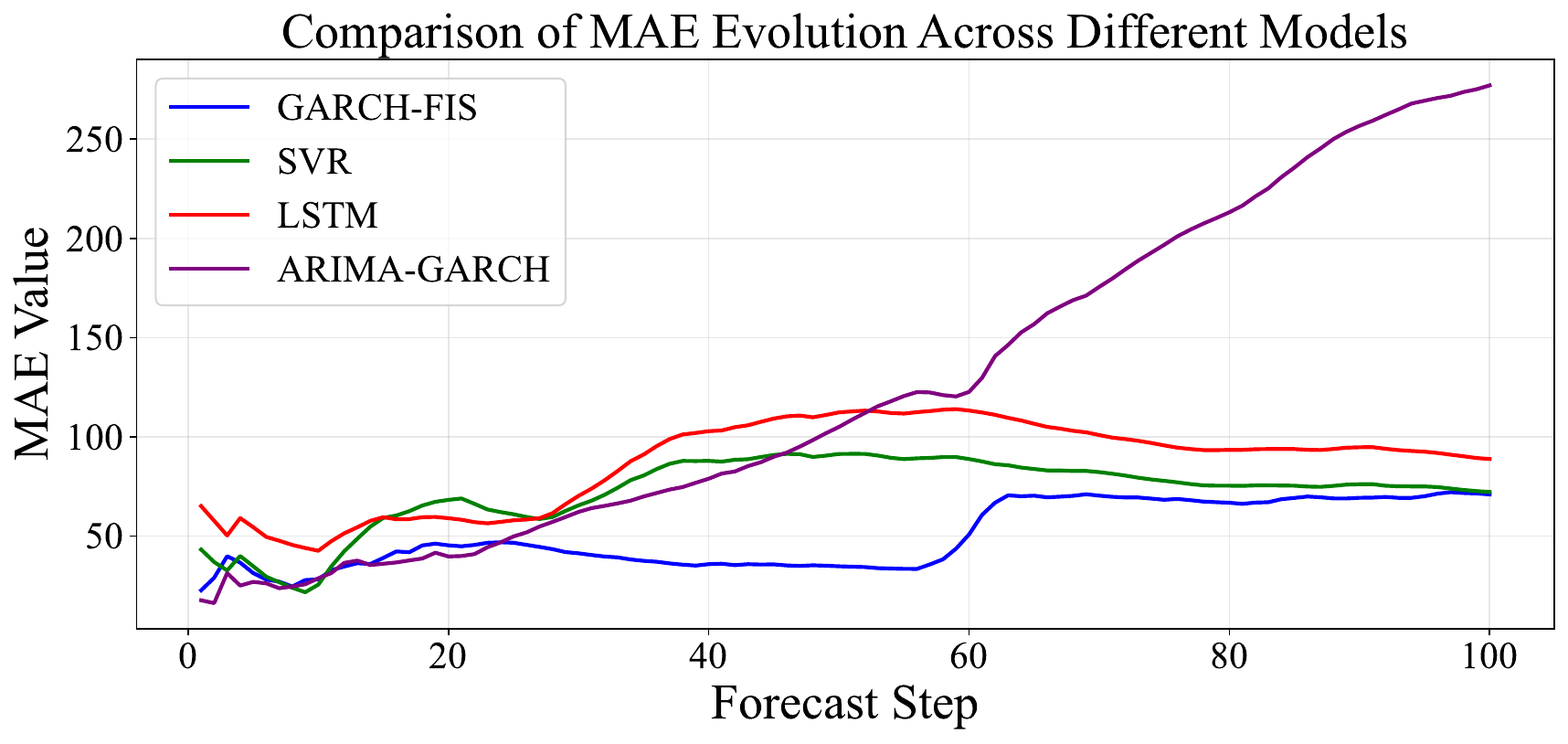}
	\subcaption{CSI 300 Index Dataset}
	\label{fig:5a} 
\end{subfigure}
\hfill
\begin{subfigure}[b]{0.48\textwidth}
	\centering
	\includegraphics[width=\textwidth]{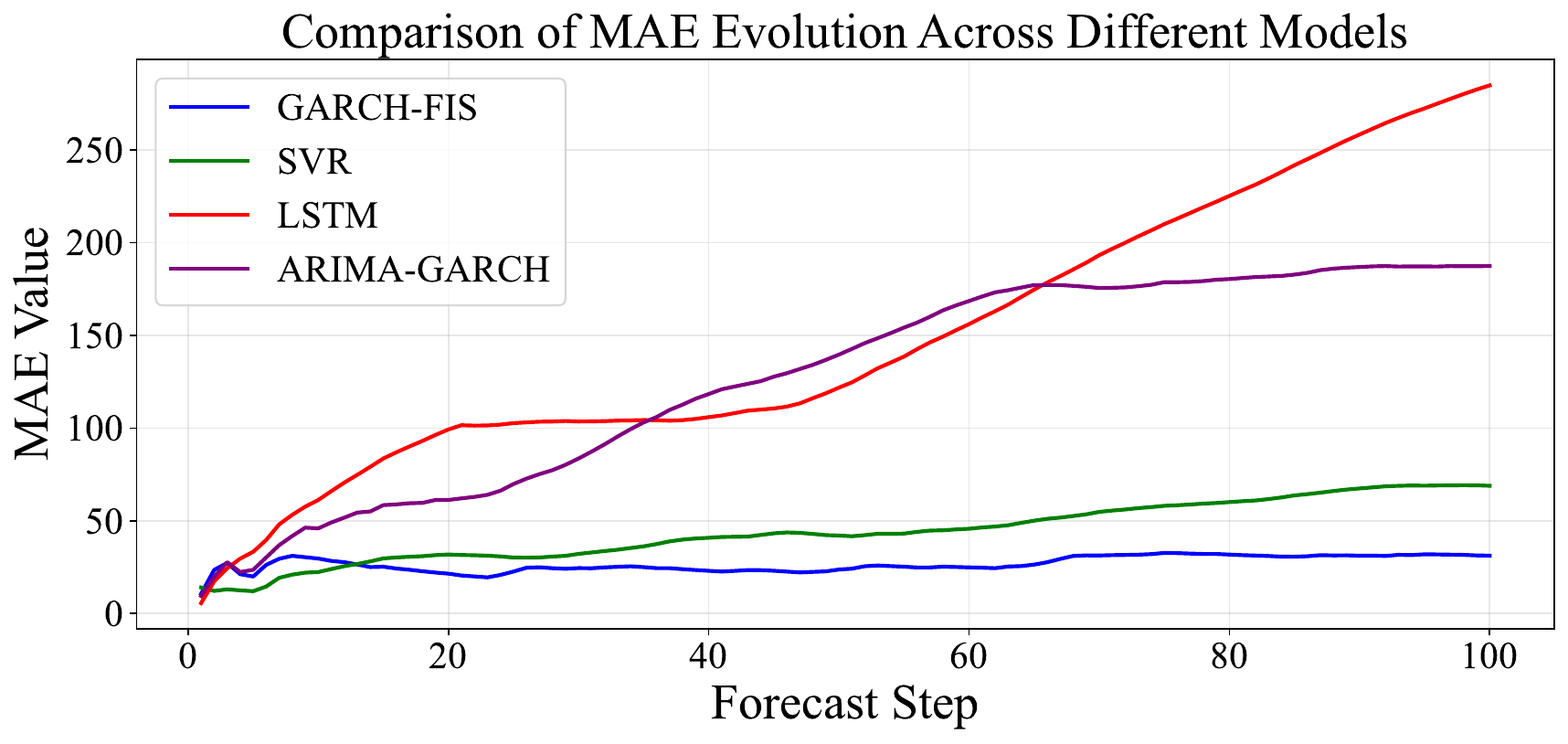}
	\subcaption{Gold Futures Dataset} 
	\label{fig:5b} 
\end{subfigure}
\vspace{0.5cm}
\begin{subfigure}[b]{0.48\textwidth}
		\centering
		\includegraphics[width=\textwidth]{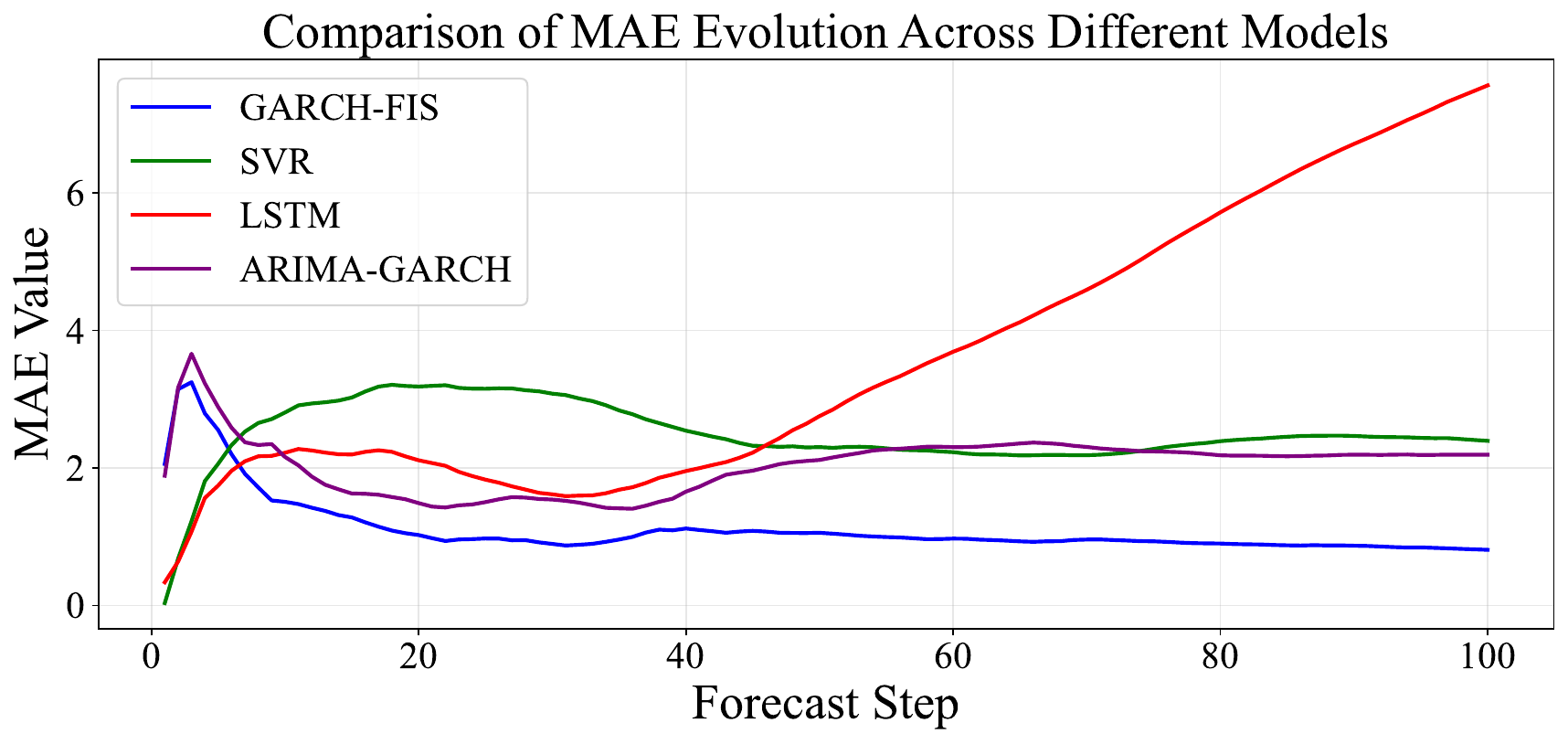}
		\subcaption{HSBC Fund Dataset}
		\label{fig:5c} 
	\end{subfigure}
	\hfill
	\begin{subfigure}[b]{0.48\textwidth}
		\centering
		\includegraphics[width=\textwidth]{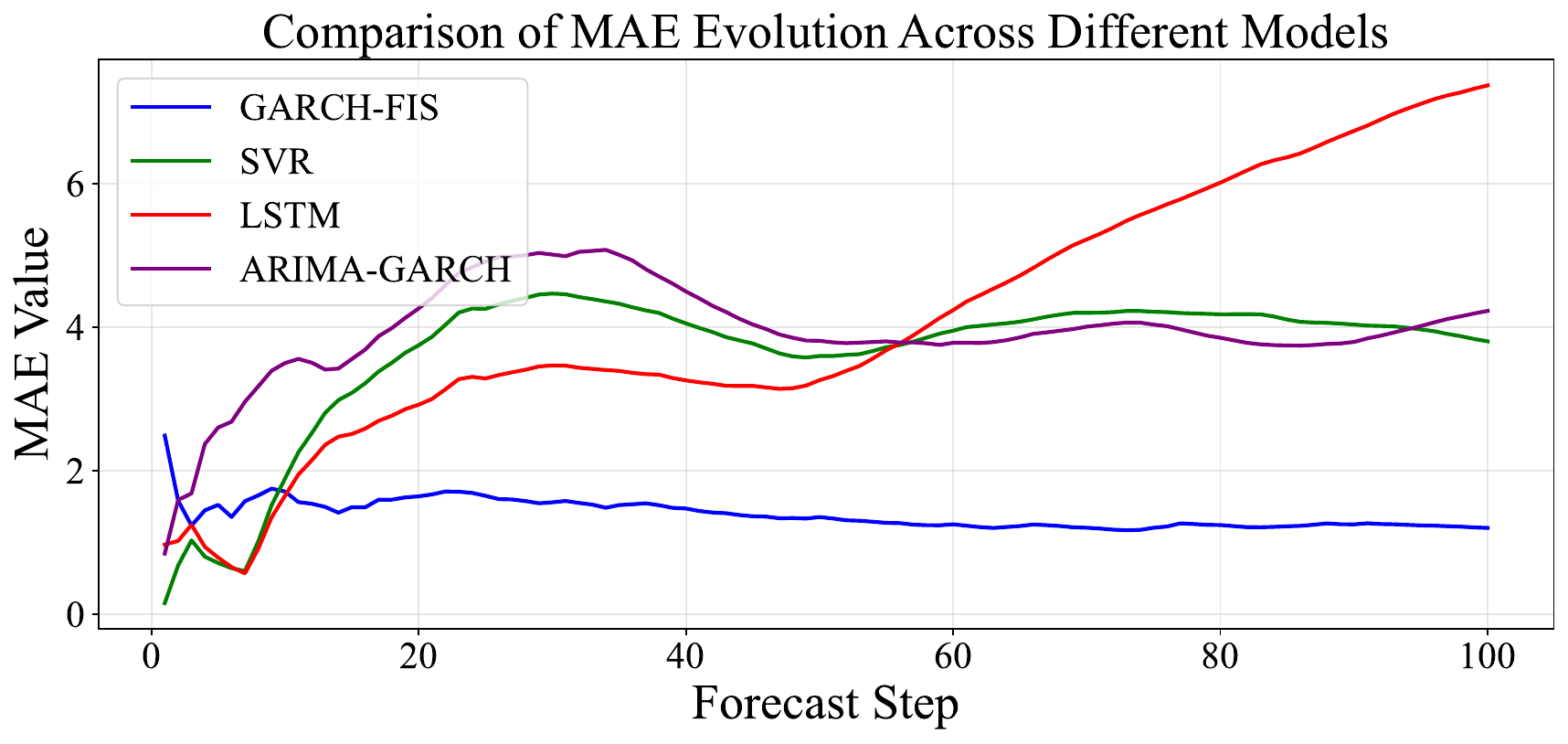}
		\subcaption{USD/JPY Exchange Rate Dataset} 
		\label{fig:5d} 
	\end{subfigure}
	\caption{MAE Trend Chart.}\label{fig:5}
\end{figure}

To visualize the error evolution and accumulation effect during multi-step forecasting, Figure \ref{fig:5} plots the trend of the MAE over the first 100 recursive prediction steps for four representative assets.

A key observation from Figure \ref{fig:5} is the consistently low and stable MAE trajectory of the GARCH-FIS model (blue solid line) across all assets. Its error shows minimal accumulation as the forecast horizon extends, confirming the robustness of the recursive rolling mechanism. In contrast, the ARIMA-GARCH model exhibits the highest and fastest-growing MAE, revealing its inability to handle non-stationarity in long-horizon forecasting. While the LSTM model performs better than ARIMA-GARCH, its MAE is significantly higher than GARCH-FIS and shows noticeable error surges at intermediate steps (e.g., 40-60 steps), reflecting the sensitivity of pure deep learning models to noise in long-sequence dependency propagation. The SVR model’s performance lies between LSTM and ARIMA-GARCH but remains inferior to the proposed method.

In summary, the GARCH-FIS model demonstrates a decisive advantage in both one-step and multi-step forecasting. It achieves the lowest error metrics, possesses the strongest explanatory power, and exhibits remarkable stability against error accumulation, validating the effectiveness of its integrated design across diverse financial time series.

\subsection{Case Study and Visualization}

\begin{figure}
 \centering
 \includegraphics[width=\textwidth]{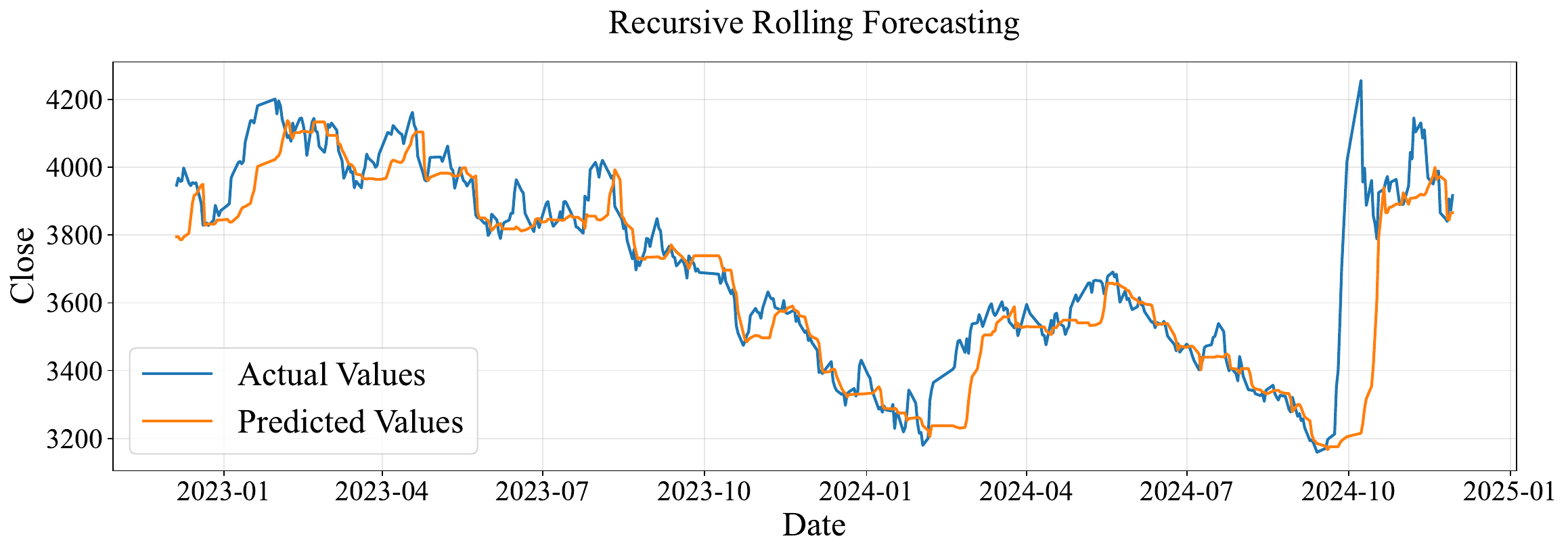}
 \caption{CSI 300 Index Prediction Comparison Chart.} 
 \label{fig:6} 
\end{figure}

To provide an intuitive validation of the model's forecasting effectiveness, we present a detailed visual analysis using the CSI 300 Index as a representative case. Figure \ref{fig:6}  compares the actual closing prices with the predicted values from the GARCH-FIS model over a segment of the test period.

Figure \ref{fig:6} illustrates that, within the forecast period from 2023 to 2024, the model's predictions (red line) closely track the actual price movements (blue line), which primarily fluctuated between 3,200 and 4,200. Specifically, during the oscillating phase in early to mid-2023, the predicted values effectively capture the turning points and amplitude of the actual fluctuations. Although minor deviations exist in local details, such as in late 2024, the predictions successfully replicate the overall trend direction and major price movements.

These results visually confirm that the GARCH-FIS model, operating within its recursive rolling framework, is capable of effectively capturing the underlying dynamics and trend patterns of financial prices. The close alignment between predicted and actual paths demonstrates the model's practical utility for financial asset price forecasting.

\subsection{Discussion}

The experimental results establish the superior performance of the GARCH-FIS model across diverse financial time series. This performance advantage can be attributed to three interconnected core mechanisms within its design.

First, the volatility-adaptive parameter adjustment mechanism is fundamental. The conditional volatility estimated by the GARCH(1,1) model is converted into price volatility (\(\hat{\sigma}_t\)) and directly regulates the width of the triangular membership functions in the FIS. 
 
Correspondingly, the width of the membership functions scaled by the same proportion, validating that the fuzzy system successfully widens for robustness during turmoil and narrows for precision during calm periods.

Second, the rolling window recursive forecasting framework ensures timeliness and combats error accumulation. The use of a 10-period rolling window allows the model to continuously integrate the latest market information, avoiding biases from stale data. The recursive prediction strategy further enables stable multi-step forecasting. The results in Figure 4 clearly show that the MAE of GARCH-FIS remains low and stable over 100 prediction steps, while the errors of traditional models escalate significantly, demonstrating this framework's effectiveness for long-horizon tasks.

Third, the intrinsic nonlinear modeling capability of the FIS effectively captures complex market patterns. Financial prices are influenced by nonlinear factors such as leverage effects, asymmetric policy impacts, and sudden sentiment shifts. Through its differentiable membership functions and rule-based inference, the FIS naturally characterizes these relationships without requiring explicit, complex parametric forms. The model's consistent success across stocks, commodities, forex, and bonds underscores its strong generalization capability for varied financial series.

In summary, the GARCH-FIS model synthesizes the dynamic volatility estimation of GARCH, the  adaptation of a rolling window, and the nonlinear mapping of a fuzzy system into a coherent and efficient forecasting framework. This synthesis provides an effective technical pathway for forecasting in complex, non-stationary financial markets.

\section{Conclusion}

This study proposed the GARCH-FIS model, a hybrid forecasting framework that integrates a FIS with a GARCH model to capture both the nonlinear dynamics and time-varying volatility of financial time series. The model's effectiveness is achieved through three key mechanisms: (1) employing the FIS for nonlinear mapping with an automatically generated rule base via the WM method, ensuring both representational power and interpretability; (2) utilizing the GARCH(1,1) model to estimate conditional volatility, which is converted into a price-volatility measure to dynamically adjust the width of the FIS membership functions, enabling adaptation to different market regimes; and (3) implementing a rolling window recursive forecasting procedure that ensures real-time adaptability and robustness in multi-step prediction. Empirical results across diverse financial assets demonstrate that the proposed model significantly outperforms benchmark methods, including SVR, LSTM, and ARIMA-GARCH, in terms of forecast accuracy and stability.

The current work has two main limitations. First, the model is designed for single-asset price forecasting and does not account for cross-asset spillover effects, limiting its direct application to portfolio decisions. Second, the input features are primarily historical prices, omitting other potential drivers such as macroeconomic indicators or market sentiment.

Future research will focus on:  exploring the deep integration of deep learning architectures with the FIS to enhance nonlinear modeling capacity and  investigating the incorporation of multivariate GARCH models within the hybrid framework to capture cross-asset volatility dependencies.

\singlespacing

%

\end{document}